\documentstyle[12pt,aclap]{article}

\newcommand{\bb}{\begin{itemize} }
\newcommand{\compact}{\setlength{\itemsep}{-.005in}}
\newcommand{\cb}{\begin{itemize} \compact}

\newcommand{\eb}{\end{itemize}}

\normalsize
\begin{document}
\title{Applying Reliability Metrics to Co-Reference Annotation}
 
\author{Rebecca J. Passonneau \\
Columbia University\\
Department of Computer Science \\
Technical Report CUCS-017-97\\
\today\\
\begin{minipage}[h]{6in}\small
\begin{center}
\vspace*{.25in}
{\large \bf Abstract}
\end{center}
Studies of the contextual and linguistic factors that
constrain discourse phenomena such as reference are coming to depend
increasingly on annotated language corpora.  In preparing the corpora,
it is important to evaluate the reliability of the annotation, but
methods for doing so have not been readily available.  In this report,
I present a method for computing reliability of coreference
annotation.  First I review a method for applying the information
retrieval metrics of recall and precision to coreference annotation
proposed by Marc Vilain and his collaborators.  I show how this method
makes it possible to construct contingency tables for computing
Cohen's $\kappa$, a familiar reliability metric.  By comparing recall
and precision to reliability on the same data sets, I also show that
recall and precision can be misleadingly high.  Because $\kappa$
factors out chance agreement among coders, it is a preferable measure
for developing annotated corpora where no pre-existing target annotation
exists.
\end{minipage}
}
 
\maketitle
\vspace*{1.3in}

\section{Two Reliability Metrics}
\label{2metrics}

{\large Two equivalent metrics for quantifying inter-rater reliability
between pairs of coders are Cohen's $\kappa$ coefficient of
agreement~\shortcite{cohen60} and Krippendorff's
$\alpha$~\shortcite{krippendorff80}.  The formulas for each are shown
in (\ref{kappa}) and (\ref{alpha}).

{\small
\begin{eqnarray}
\label{kappa} \kappa &=& \frac{p_{A_{O}}-p_{A_{E}}}{1-p_{A_{E}}} \\
\label{alpha} \alpha &=& 1 - \frac{p_{D_{O}}}{p_{D_{E}}} \\
\nonumber & & \\
\label{obs} 1 &=& p_{A_{O}} + p_{D_{O}} \\
\label{exp} 1 &=& p_{A_{E}} + p_{D_{E}} \\
\nonumber & & \\
\label{deriv1} \frac{p_{A_{O}}-p_{A_{E}}}{1-p_{A_{E}}} &=& 
   1 -  \frac{p_{D_{O}}}{p_{D_{E}}} \\
\label{deriv2} \frac{p_{A_{O}}-p_{A_{E}}}{1-p_{A_{E}}}  &=& 
   1 -  \frac{(1-p_{A_{O}})}{(1-p_{A_{E}})} \\
\label{deriv3} \frac{p_{A_{O}}-p_{A_{E}}}{1-p_{A_{E}}}  &=& 
  \frac{(1-p_{A_{E}})}{(1-p_{A_{E}})} - \frac{(1-p_{A_{O}})}{(1-p_{A_{E}})} \\
\label{deriv5} \frac{p_{A_{O}}-p_{A_{E}}}{1-p_{A_{E}}}  &=& 
  \frac{p_{A_{O}} - p_{A_{E}}}{(1-p_{A_{E}})} 
\end{eqnarray}
}

\vspace*{1.3in}
Briefly, Cohen's $\kappa$ is cast in terms of the amount of {\it
agreement} between coders that exceeds chance expectations.  The
numerator of the ratio in (\ref{kappa}) is the proportion of observed
agreements ($p_{A_{O}}$) less the proportion expected to agree by chance
($p_{A_{E}}$); the denominator is the total proportion (100\%) less the
the proportion expected to agree by chance.  Conversely, Krippendorff's
$\alpha$ is cast in terms of the extent to which the observed {\it
disagreements} between coders is below chance expectation; it is the
total probability less the ratio of observed disagreements to expected
disagreements.  The observed probability of agreement and disagreement
must sum to one, as must the expected probability of agreement and
disagreement ((\ref{obs}) and (\ref{exp})).  By substitution, it can be
shown that $\kappa$ equals $\alpha$ ((\ref{deriv1}) - (\ref{deriv5})).

{\normalsize
\begin{table}
\begin{center}
\begin{tabular}{c r r r }
& \multicolumn{2}{c}{Judge Y} & \\
\multicolumn{1}{c}{Judge X} &
   \multicolumn{1}{c}{A} & \multicolumn{1}{c}{B}  & \\\cline{2-3}
A &\multicolumn{1}{|r}{47}  & \multicolumn{1}{|r|}{14}  &  {\it 61} \\\cline{2-3}
B &\multicolumn{1}{|r}{10}  & \multicolumn{1}{|r|}{29}  &  {\it 39}\\\cline{2-3}
  & {\it 57} & {\it 43} & {\it 100}\\
\end{tabular}
\begin{eqnarray}
\label{reli2by2} \alpha & = \kappa & = .50 
\end{eqnarray}
\end{center}
\caption{\label{2by2coin} A 2-by-2 coincidence matrix}
\end{table}
}

The reliability measures depend crucially on a hypothesis of chance
expectation.  In~\cite{cohen60} and ~\cite{krippendorff80}, chance
expectation is derived from the marginals of a coincidence matrix
classifying the response categories of one coder by the response
categories of another coder.  Table~\ref{2by2coin} illustrates a
simple 2-by-2 coincidence matrix.  A coincidence matrix classifies a
set of data in a way that shows, for a given set of classification
categories (e.g., A versus B), how the data is cross-classified.
Every data point must go in one and only one cell of the table to
indicate how the data classified by one coding (row categories) is
cross-classified by the other coding (column categories).  The
diagonal from upper left to lower right in Table~\ref{2by2coin}
represents the responses of judge X that {\it coincide} with judge
Y's; cells off the diagonal represent classification disagreements.

The marginals in Table~\ref{2by2coin} show that 61\% of judge X's
responses are in category A compared with 57\% of Y's.  Where .61 is
taken to be the likelihood that X responds in category A, and .57 the
likelihood that Y responds in category A, then .61 $\times$ .57 of the
time X and Y should agree that the same data point is classified in
category A, assuming nothing more than chance correspondence between X
and Y's responses.  Adding the result of the corresponding likelihood
of agreement on response B yields $p_{A_{E}}$ = 52\%.  The expected
proportion of disagreement is similarly computed.  By chance, X should
respond A where Y responds B 26\% of the time (.61 $\times$ .43). The
difference between these expected values and the observed agreements
(.47 + .29) results in a reliability value of .50, as shown in
(\ref{reli2by2}) of Table~\ref{2by2coin}.

Whenever the responses of two subjects can be cast in the form of a
coincidence matrix, the reliability metrics illustrated above can be
applied.  Here I present a proposal for applying reliability to
coreference annotation, based on the insights in ~\cite{vilain&etal95}.

}

\section{Evaluating Coreference Annotations}
\label{prerequ}

{\large Co-reference annotation is annotation of language data to
indicate when distinct expressions have been used to corefer.
Evaluating the reliability of such data is important for several
reasons.  First, any annotation task is subject to unintended errors
arising from lack of attention on the part of the annotator.  The
likelihood of such errors depends in part on ergonomic factors such as
what kinds of aids are provided for recording and checking annotations,
and how much time the annotator has to perform the task.  In addition, no
matter how precise a language user might be, language interpretation is
subjective.  A given expression can be referentially ambiguous or vague.
Referential indeterminacy can even be intentional on the part of the
speaker or writer. When annotations of the same data are collected from
two or more coders, then in principle, the reliability of the data (or of
the individual coders) can be quantified.

{\small
\begin{figure*}[t]
\begin{tabular}{l l}
\multicolumn{1}{c}{Sample 1: Journalistic Text}  &
\multicolumn{1}{c}{Sample 2: Problem-Solving Dialogue} \\
\begin{minipage}[h]{3in}
\hspace{.25in}Committee approval of $[$Gov. Price Daniel's {\it abandoned property}
act$]_{1}$ seemed certain Thursday despite the adamant protests of Texas
bankers.  $[$Daniel$]_{2}$ personally led the fight for $[$the
measure$]_{1}$, which $[$he$]_{2}$ had watered down considerably since its
rejection by two previous Legislatures, in a public hearing before the
House Committee on Revenue and Taxation.  Under committee rules,
$[$it$]_{1}$ went automatically to a subcommittee for one week. But
questions with which $[$committee members$]_{3}$ taunted bankers
appearing as witnesses left little doubt that $[$they$]_{3}$ will
recommend passage of $[$it$]_{1}$.
\end{minipage} 
&
\begin{minipage}[h]{3in}
M: okay we need to ship a boxcar of oranges to Bath by 8 AM today S:
okay M: umm okay so I guess uh I would suggest that we use $[$engine
E1$]_{1}$ uh and have $[$it$]_{1}$ pick up $[$a boxcar$]_{2}$ at ah
Dansville how long'll $[$it$]_{1}$ take S: uh that'll take 3
hours to get to Dansville and get $[$the boxcar$]_{2}$ M: uh okay and
then how long to go on to .. Corning with $[$the boxcar$]_{2}$ coupled
to uh $[$E1$]_{1}$ S: another hour M: ok so that's okay and then uh if
we loaded $[$the oranges$]_{3}$ at ah Corning and sent ah $[$E1$]_{1}$
on to Bath with $[$the oranges$]_{3}$ S: we'd get there at 7
\end{minipage} \\ 
& \\\hline
\multicolumn{2}{c}{Co-reference Annotations} \\\hline
\begin{minipage}[h]{3in}
\begin{tabular}{llr}
\multicolumn{1}{c}{Token} &
\multicolumn{1}{c}{String} &
\multicolumn{1}{c}{CA} \\\hline
A & Gov. Price Daniel's \ldots act & 1	\\
B & Daniel	& 2	\\
C & the measure	& 1	\\
D & he		& 2	\\
E & it		& 1	\\
F & comittee members	& 3	\\
G & they	& 3	\\
H & it		& 1	\\
 & \\
 & \\
\end{tabular}
\end{minipage} &
\begin{minipage}[h]{3in}
\begin{tabular}{llrr}
\multicolumn{1}{c}{Token} &
\multicolumn{1}{c}{String} &
\multicolumn{1}{c}{CA$_{1}$} & \multicolumn{1}{c}{CA$_{2}$} \\\hline
A & engine E1	& 1 & 1	\\
B & it		& 1 & 1	\\
C & a boxcar	& 2 & 2	\\
D & it		& 1 & 2	\\
E & the boxcar	& 2 & 2	\\
F & the boxcar	& 2 & 2	\\
G & E1		& 1 & 1	\\
H & the oranges	& 3 & 3	\\
I & E1		& 1 & 1	\\
J & the oranges	& 3 & 3	\\
\end{tabular}
\end{minipage}\\ \hline
\end{tabular} 
\caption{\label{coref-annot} Co-reference annotation of two language samples} 
\end{figure*}
}

Two language samples are presented in Figure~\ref{coref-annot} that
typify two quite different types of discourse.  Sample 1 illustrates
journalistic text, and is taken from the Brown
Corpus~\cite{francis&kucera82}.  Sample 2, illustrating spoken
dialogue, is from the University of Rochester's Trains 91
corpus~\cite{trains91}.  Two samples are shown to illustrate that
despite major differences of language variety, the task of coreference
annotation is essentially the same for both types of data.  Both
samples have been annotated to indicate certain expressions that have
been interpreted to corefer (how or why these particular expressions
were selected is immaterial to the present discussion).  Relevant
phrases have been bracketed.  Bracketed phrases that have been
annotated with the same numeric subscript represent expressions that,
in the annotator's judgement, were used to corefer.  For sample 1,
eight expressions (A-H) were annotated as referring to one of three
distinct referents.  The coding of co-referential expressions is shown
under column CA (Coreference Annotation).  For sample 2, ten
expressions (A-J) were annotated as referring to one of three distinct
referents, whose indices are listed under the column headed CA$_{1}$.
An alternate coding is shown in column CA$_{2}$.  The remainder of the
discussion will focus on sample 2.

How can a comparison of the two annotations of sample 2, CA$_{1}$ and
CA$_{2}$, be quantified?  The key observations used
in~\cite{vilain&etal95} are that the sets of expressions that corefer
constitute equivalence classes, and that in two annotations, a given
expression is either assigned to the same equivalence class or not.  I
first present how~\cite{vilain&etal95} compute precision and recall by
comparing equivalence classes across a pair of annotations.  Then I
show how a revision of their approach can be converted to reliability
measures, under certain important constraints.

The first annotation for Sample 2 places five tokens into one
equivalence class referring to the engine (\{A, B, D, G, I\}), and
three tokens into a class referring to the boxcar (\{C, E, F\}).  This
contrasts with the alternate annotation, where the same eight tokens
are in two equivalence classes, but where D is placed with C: \{A, B,
G, I\}, \{C, D, E, F\}.  To apply recall and precision, we must assume
that one of the annotations is {\it correct}.  In general, a recall
error involves failure to identify members of a target set; a
precision error involves inclusion of additional elements besides
those in the target set.  Vilain et al.~\shortcite{vilain&etal95}
observe that intuitively, a comparison of two sets \{A, B, D, G, I\}
from CA$_{1}$ and \{A, B, G, I\} from CA$_{2}$, where the first set is
the target, involves only a recall error.  The CA$_{2}$ set does not
include any additional elements, but it fails to include D.  In
contrast, the comparison of \{C, E, F\} as the target with \{C, D, E,
F\} involves a precision error and no recall errors. In practice, the
method given in~\cite{vilain&etal95} does not compare elements of
corresponding sets, but compares how many links are needed to connect
the elements within corresonding sets.

To compute recall, ~\cite{vilain&etal95} start by creating
a partition of a given target set from the corresponding response sets.
This addresses the question of how many equivalence classes in the
response set must be examined in order to reconstruct the target set.
The relevant partition of \{A, B, D, G, I\} is thus into the two sets
\{A, B, G, I\}, \{D\}.  If the target set is conceived of as five nodes
in a spanning tree (e.g., A--B--D--G--I), then the target ``tree'' can be
constructed from the response by adding one link: a link from D to any
node A, B, G or I.  In general, the missing information for recall is
quantified in terms of the number of links missing from the response
partition.  The number of links in a target equivalence class C is the
cardinality of that class less 1: $|C| \: - \: 1$.  The number of links
missing from the partition of C relative to the response ($p(C)$) is the
cardinality of the partition less 1: $|p(C)| \: - \: 1$.  The recall for
a given equivalence class is thus the ratio of the target links less the
missing links to the target links:

\begin{eqnarray}
\label{recall1}
\nonumber 
Recall_{C} & = & \frac{(|C| \: - \: 1) \: - \: (|p(C)| \: - \: 1)}{(|C| \: - \: 1)}\\
\nonumber & & \\
& = & \frac{|C| \: - \: |p(C)|}{(|C| \: - \: 1)}
\end{eqnarray}

When an equivalence class $C_{i}$ in the target has an exact
correspondence to one in the response, the cardinality of the partition
$p(C_{i})$ is 1, the numerator and denominator in (\ref{recall1}) are
the same, and recall is perfect.  Recall for a complete annotation is
expressed in terms of all the equivalences classes $C_{i}$ in the target
annotation, by summing the recall errors (numerator) and summing the
target links (denominator):

\begin{eqnarray}
\label{recall2}
Recall = &  \frac{\sum_{i} |C_{i}| \: - \: |p(C_{i})|}{\sum_{i} (|C_{i}| \: - \: 1)} \\
\label{recall2a} Recall_{CA_{1},CA_{2}} & \frac{(5-2)+(3-1)+(2-1)}{(5-1)+(3-1)+(2-1)} 
\end{eqnarray}

Taking $CA_{1}$ as the target, formula (\ref{recall2}) gives a recall
for $CA_{2}$ of .86, as shown in (\ref{recall2a}).

Computation of precision in~\cite{vilain&etal95} is the
converse of the computation of recall.  To illustrate, precision will be
computed for the target set \{C, E, F\}.  Precision is imperfect because
the response set has an additional member: \{C, D, E, F\}.  Where the
{\it response} set is R, a partition of the response set relative to the
target sets ($p(R)$) gives the two sets \{C, E, F\} and \{D\}.
Precision of the {\it target} set C is then the ratio of the
difference between the cardinality of the corresponding response set R and the
cardinality of its partition p(R) to the cardinality of the response set R less
1:

{\small
\begin{figure*}[t]
M: okay we need to ship a boxcar of oranges to Bath by 8 AM today S:
okay M: umm okay so I guess uh I would suggest that we use $[$engine
E1$]_{1}$ uh and have $[$it$]_{1}$ pick up $[$a boxcar$]_{2}$ at ah
Dansville how long'll $[$it$]_{3}$ take S: uh $[$that$]_{3}$'ll take 3
hours to get to Dansville and get $[$the boxcar$]_{2}$ M: uh okay and
then how long to go on to .. Corning with $[$the boxcar$]_{2}$ coupled
to uh $[$E1$]_{1}$ S: another hour M: ok so that's okay and then uh if
we loaded $[$the oranges$]_{4}$ at ah Corning and sent ah $[$E1$]_{1}$
on to Bath with $[$the oranges$]_{4}$ S: we'd get there at 7
\vspace{.25in}

\begin{tabular}{ll}\hline
\begin{minipage}[h]{2.7in}
\begin{tabular}{llrr}
\multicolumn{1}{c}{Token} &
\multicolumn{1}{c}{String} &
\multicolumn{1}{c}{CA$_{1}$} & \multicolumn{1}{c}{CA$_{3}$} \\\hline
A  & engine E1	 & 1 & 1	\\
B  & it		 & 1 & 1	\\
C  & a boxcar	 & 2 & 2	\\
D  & it		 & 1 & 3	\\
D' & that        & 4 & 3 \\
E  & the boxcar	 & 2 & 2	\\
F  & the boxcar	 & 2 & 2	\\
G  & E1		 & 1 & 1	\\
H  & the oranges & 3 & 4	\\
I  & E1		 & 1 & 1	\\
J  & the oranges & 3 & 4	\\
\end{tabular}
\end{minipage}
&
\begin{minipage}[h]{2.7in}
\begin{tabular}{ll}
\multicolumn{1}{c}{CA$_{1}$ Equivalence classes} & 
\multicolumn{1}{c}{CA$_{3}$ Equvalence classes} \\\hline
\{A, B, D, G, I\} & \{A, B, G, I\} \\
\{C, E, F\} & \{C, E, F\} \\
\{H, J\} & \{D, D'\} \\
 & \{H, J\} \\
& \\
& \\
& \\
& \\
& \\
& \\
& \\
\end{tabular}
\end{minipage}\\ \hline
\end{tabular} 
\caption{\label{coref-annot2} Alternate co-reference annotation of sample 2} 
\end{figure*}
}

\begin{eqnarray}
\label{precision}
Precision_{C} = & \frac{|R| \: - \: |p(R)|}{(|R| \: -
\: 1)} \\
\nonumber & \\
\label{precision2} Precision = & \frac{\sum_{i} |R_{i}| \: - \:
|p(R_{i})|}{\sum_{i} (|R_{i}| \: - \: 1)} \\
\label{precision2a} Precision_{CA_{1},CA_{2}} = & \frac{(4-1)+(4-2)+(2-1)}{(4-1)+(4-1)+(2-1)}
\end{eqnarray}

Precision for the equivalence class \{C, E, F\} is then
$\frac{4-2}{4-1}$, or .33.  Precision of the entire coding $CA_{2}$
relative to $CA_{1}$ is .86, as shown in (\ref{precision2a}).}

\subsection{Problems}
{\large

A perhaps more realistic alternate coding for sample 2 is shown in
Figure~\ref{coref-annot2}.  The token identified in
Figures~\ref{coref-annot}-\ref{coref-annot2} as D was coded as
coreferential with the expression {\it engine E1} (token A) in
annotation $CA_{1}$.  In annotation $CA_{3}$ shown in
Figure~\ref{coref-annot2}, this token is interpreted to refer to the
process of getting engine E1 to pick up a boxcar at Dansville, and is
annotated as coreferential with a token of the demonstrative pronoun
{\it that}---shown here as token D'. D' was not originally included in
$CA_{1}$, but is given here an arbitrary index of 4 in coding $CA_{1}$
to indicate lack of coreference with any other expression.  I will use
a comparison of codings $CA_{1}$ and $CA_{3}$ to illustrate how the
approach taken in~\cite{vilain&etal95} presents certain problems for
computing reliability, and for evaluating the type of annotation
employed in~\cite{passonneau&litman97}.

Both of the problems discussed here pertain to the manner in which
recall and precision is applied to data, rather than to the actual
computation of recall and precision.  The first problem is
that~\cite{vilain&etal95} do not constrain the sets of referring
expressions that are being compared to have the same cardinality.  The
second is that they apply their method only to referring expressions
that corefer with at least one other expression.  My proposed solution
requires that two annotations have the same cardinality of referring
expressions.  It also permits an annotator to interpret an expression
as having no coreferential expressions, as in D' for coding $CA_{1}$
(Figure~\ref{coref-annot2}).  As I show below, these two moves make it
possible to retain the basic insight from~\cite{vilain&etal95}, to
compute reliability, and to apply the method to a broader range of
annotation approaches, including the annotation style presented
in~\cite{drama97}.

The fundamental problem in comparing codings $CA_{1}$ and $CA_{3}$ is
that the two data sets are incommensurate.  Coding $CA_{1}$ originally
placed ten expressions into equivalence classes, while coding $CA_{3}$
does so for eleven expressions.  This prevents creation of a
contingency table, and is thus an obstacle to applying reliability
measures (cf. section~\ref{2metrics}).

The approach in~\cite{vilain&etal95} does not require two codings to
be commensurate in part because the annotators' task, as described
in~\cite{muccs4.0}, has two parts: to identify the expressions to be
coded, or {\it markables}, and to place markables into equivalence
classes based on the coreference relation.  As I argue
in~\cite{drama97}, there are several disadvantages to this approach.
Identifying markables is a conceptually distinct task, can be partly
automated with easily accessible and relatively simple tools, such as
part-of-speech taggers, and is a language specific task.  In contrast,
coreference is difficult to automate (particularly in a sufficiently
general way to apply across corpora), and is language independent.  I
take the evaluation of how markables are identified to be a separate
problem.  My goal is then to evaluate the inter-rater reliability of
co-reference annotations, assuming that each rater is given the same
set of markables to annotate.

Another serious drawback, of particular concern to investigators in
the natural language generation community, is that the approach taken
in~\cite{vilain&etal95} fails to identify referential expressions
comprising a singleton equivalence class.  Instead, such expressions
are omitted from consideration.  However, it is of as much concern to
determine the conditions under which a referent is mentioned only
once, as to determine those under which it is re-mentioned.  If two
coders place the same expression in a class by itself, indicating lack
of any coreferential expressions, note that recall and precision will
both be zero.  While at first this may seem counter-intuitive, it is
entirely reasonable.  First, what is being evaluated is the ability of
distinct coders to find the same coreference links.  In the case of
comparing a singleton set to an identical singleton set, there are no
coreference links to find.  But note that no mismatching links have
been identified.

{\small
\begin{figure}[h]
\begin{center}
\begin{tabular}{c r r r }
& \multicolumn{2}{c}{Coding $CA_{1}$} & \\
\multicolumn{1}{c}{Coding $CA_{3}$} &
   \multicolumn{1}{c}{+Link} & \multicolumn{1}{c}{-Link}  & \\\cline{2-3}
+Link &\multicolumn{1}{|r}{a}  & \multicolumn{1}{|r|}{b}  &  a+b \\\cline{2-3}
-Link &\multicolumn{1}{|r}{c}  & \multicolumn{1}{|r|}{d}  &  c+d\\\cline{2-3}
  & a+c & b+d & a+b+c+d\\ 
\end{tabular} 
{\small
\begin{eqnarray} 
\label{newrecall}
Recall & = & \frac{a}{a+c} =
\frac{\sum_{i} |C_{i}| \: - \: |p(C_{i})|}{\sum_{i} (|C_{i}| \: - \: 1)} 
\end{eqnarray} 
\begin{eqnarray} 
\label{newprecision}
Precision & =  & \frac{a}{a+b}
    = \frac{\sum_{i} |C'_{i}| \: - \: |p(C'_{i})|}{\sum_{i} (|C'_{i}| \: - \: 1)}
\end{eqnarray} }
\end{center}
\caption{\label{2by2schematic} Schematic representation of a 2-by-2
coincidence matrix} 
\end{figure} } 

Consider the result of imposing the requirement that two coreference
codings must partition the same set of expressions into equivalence
classes of coreference.  If we assume that coding $CA_{1}$ represents
an annotator's judgement that token D' is in a singleton set, then we
can create a contingency table of the two codings. The table total
represents the total number of possible coreference links.  In the
case of codings $CA_{1}$ and $CA_{3}$, the table total is the
cardinality of the set of tokens less 1, which is ten.  To compute
reliability, we need the four quantities $a \: - \: d$ given in each
cell of the table shown in Figure~\ref{2by2schematic}
(cf. Table~\ref{2by2coin}).  Of all possible coreference links, some
will be identified by both coders.  This is quantity $a$ in
Figure~\ref{2by2schematic}.  Some will be identified by neither coder:
quantity $d$ in Figure~\ref{2by2schematic}.  Thus $a$ and $d$
represent the two types of agreement between coders: agreement on
coreference links, and agreement on their absence.  In contrast,
quantities $b$ and $c$ represent disagreements: the first coder finds
links that the second coder does not, or vice versa.

Recall and precision are defined as illustrated in (\ref{newrecall})
and (\ref{newprecision}) of
Figure~\ref{2by2schematic}~\cite{vanRijsbergen79}.  Recall represents
the ratio of links found in both the target and some test set, hence
is the ratio of $a$ to $a\: + \: c$.  By setting this ratio equal to
(\ref{recall1}), the ratio proposed in~\cite{vilain&etal95}, we can
begin to identify the individual quantities $a$ through $d$.
Precision represents the proportion of links found in some test set
that are also in the target, hence is the ratio of $a$ to $a\: + \:
b$.  As shown in (\ref{newprecision}), this ratio can be equated to
(\ref{precision2}).  Given the table total and the two equalities
(\ref{newrecall}) and (\ref{newprecision}), the four quantities $a$
through $d$ can be computed.

Recall that quantity $a$ is the coreference links agreed on by
$CA_{1}$ and $CA_{3}$.  By (\ref{newrecall}) and (\ref{newprecision}),
it is the sum of the differences of the cardinality of each
equivalence class in $CA_{1}$ less the cardinality of its partition by
$CA_{3}$. Equivalently, $a$ is the sum of the differences of the
cardinality of each equivalence class in $CA_{3}$ less the cardinality
of its partition by corresponding equivalence classes in $CA_{1}$:

{\small
\begin{eqnarray*}
\label{compute-a}
\nonumber a  =& (5-2) + (3-1) + (1-1) + (2-1) \\
\nonumber & \\
\nonumber a  =& (4-1) + (3-1) + (2-2) + (2-1) \\
\nonumber & \\
a  =& 6 
\end{eqnarray*}}

\vspace{-.1in}
Cell value $b$ represents the coreference links identified in $CA_{3}$
but not in $CA_{1}$.  It is the sum of the number of links for each
equivalence class in $CA_{3}$ ($\sum_{i} \: |C_{i}| -1$) less the
coreference links found by both:

{\small
\begin{eqnarray*}
\label{compute-b}
\nonumber b  =& ((4-1) + (3-1) + (2-1) + (2-1)) - 6 \\
b  =& 1\\
\end{eqnarray*}}

\vspace{-.1in}
Conversely, cell value $c$ represents the coreference links identified
in $CA_{1}$ but not in $CA_{3}$.  It is the sum of the number of links
for each equivalence class in $CA_{1}$ less the coreference links
found by both:

{\small
\begin{eqnarray*}
\label{compute-c}
\nonumber c  =& ((5-1) + (3-1) + (1-1) + (2-1)) - 6 \\
c  =& 1\\
\end{eqnarray*}}

\vspace{-.1in} It remains to calculate $d$, the possible links that
neither coder identifies.  We know the total possible coreference
links: $a \: + \: b \: + \: c +\: d\: =\: 10$.  And we know the values
of a, b and c (a=6; b=c=1), thus d = 2.  Another way to compute $a$
and $d$ is to compute the full partition of the equivalence classes in
both codings (p(CA)), giving all links found in both codings:

{\small
p(CA) = \{A, B, G, I\}, \{C, E, F\}, \{D\}, \{D'\}, \{H, J\}}

Note that the value of $a$ (links agreed on by both coders) is the sum
of the differences of the cardinality of each set in the partition
p(CA) less 1:

{\small
\begin{eqnarray*}
\label{compute-a2}
\nonumber a  =& (4-1) + (3-1) + (1-1) + (1-1) + (2-1) \\
\nonumber a  =& 6
\end{eqnarray*}}

\vspace{-.1in}
Then take the intersection of either $CA_{1}$ or $CA_{3}$ with p(CA).
The value of $d$ is the cardinality of either intersection less 1:

{\small
\begin{eqnarray}
\label{intersect}
\nonumber CA_{1} \: \cap \: p(CA) & = & \{C, E, F\}, \{D'\}, \{H, J\} \\
\nonumber & & \\
\nonumber CA_{3} \: \cap \: p(CA) & = & \{A, B, G, I\}, \{C, E, F\}, \{H, J\} \\
\nonumber d & = & |CA_{1} \: \cap \: p(CA)| -1 \\
\nonumber d & = & |CA_{3} \: \cap \: p(CA)| -1 \\
\nonumber d & = & 2 
\end{eqnarray}}

\vspace{-.1in}
The contingency table for comparing CA$_{1}$ and CA$_{3}$ using the cell
values we have just computed is given in Table~\ref{c1byc3}.

{\small
\begin{table}
\begin{center}
\begin{tabular}{c r r r }
& \multicolumn{2}{c}{Coding $CA_{1}$} & \\
\multicolumn{1}{c}{Coding $CA_{3}$} &
   \multicolumn{1}{c}{+Link} & \multicolumn{1}{c}{-Link}  & \\\cline{2-3}
+Link &\multicolumn{1}{|r}{6}  & \multicolumn{1}{|r|}{1}  & 7  \\\cline{2-3}
-Link &\multicolumn{1}{|r}{1}  & \multicolumn{1}{|r|}{2}  & 3 \\\cline{2-3}
  & 7 & 3 & 10\\ 
\end{tabular} 
\begin{eqnarray} 
Recall & = \frac{6}{7} & = 85.7\% \\
Precision & = \frac{6}{7} & = 85.7\%
\end{eqnarray} 
\end{center}
\caption{\label{c1byc3} Coincidence matrix for CA$_{1}$ by CA$_{3}$}
\end{table} }

} 

\subsection{Conversion to Reliability}

{\large 

Now that we see how to construct a contingency table for coreference
annotation, it is straightforward to compute reliability.  Given that
recall and precision are both just over 85\%, one might interpret the
similarity of the coding as being moderately good.  However, as shown
in (\ref{this-reli1})-(\ref{this-reli9}), reliability is poor.  The interpretation of the
$\kappa$ value of .52 is that reliability is about halfway between
completely random behavior ($kappa$ = 0) and perfect reliability (near
1).\footnote{A negative $kappa$ value represents positive
unreliability, as opposed to random correspondence. See~\cite{cohen60}
for a discussion of the upper and lower limits of $\kappa$ assuming
p$_{A_{E}}$ is derived from marginals of a coincidence matrix.
See~\cite{krippendorff80} for other methods of computing p$_{A_{E}}$,
and for applying reliability to continuous variables, etc.}

{\small
\begin{eqnarray}
\label{this-reli1}
\kappa & = & \frac{p_{A_{O}} - p_{A_{E}}}{1-p_{A_{E}}} \\
\label{this-reli2} p_{A_{O}} & = & .6 + .2 \\
\label{this-reli3} p_{A_{E}} & = & (.7\times.7) + (.3\times.3) \\
\label{this-reli4} \kappa & = & 
  \frac{(.6+.2)-((.7\times.7)+(.3\times.3))}{1-((.7\times.7)+(.3\times.3))} \\
\label{this-reli5} \alpha & = & 1 - \frac{p_{D_{O}}}{p_{D_{E}}} \\
\label{this-reli6} p_{D_{O}} &=& .1+.1\\
\label{this-reli7} p_{D_{E}} &=& (.7\times.3) + (.7\times.3)\\
\label{this-reli8} \alpha & = & 1 -
  \frac{(.1+.1)}{((.7\times.3)+(.7\times.3))} \\
\label{this-reli9}\kappa & = & \alpha = .52
\end{eqnarray}}
}

{\small
\begin{table}
\begin{center}
\begin{tabular}{c r r r }
& \multicolumn{2}{c}{Coding $R_{1}$} & \\
\multicolumn{1}{c}{Coding $R_{2}$} &
   \multicolumn{1}{c}{+Link} & \multicolumn{1}{c}{-Link}  & \\\cline{2-3}
+Link &\multicolumn{1}{|r}{166}  & \multicolumn{1}{|r|}{19}  & 185  \\\cline{2-3}
-Link &\multicolumn{1}{|r}{13}  & \multicolumn{1}{|r|}{44}  & 57 \\\cline{2-3}
  & 179 & 63 & 242 \\ 
\end{tabular} 
\begin{eqnarray} 
Recall & = .90\% \\
Precision & = .93\%
\end{eqnarray} 
\begin{eqnarray} 
\kappa & = & .65
\end{eqnarray} 
\end{center}
\caption{\label{real-data} Coincidence matrix for R$_{1}$ by R$_{2}$}
\end{table} }

Table~\ref{real-data} compares the $\kappa$ reliability score with
recall and precision for an actual coding of a spoken narrative
from~\cite{chafe80}.  One coding represents the consensus coding of
coreference arrived at by the two investigators in the study reported
in~\cite{passonneau&litman97}.  The other coding was performed by a
student with no linguistics background but some training in
coreference annotation.  As illustrated, the recall and precision
scores are both apparently good (90\% or above), but the $\kappa$
score is only .65.  This demonstrates concretely that because recall
and precision do not factor out chance agreement, they can be
misleading. In contrast, as discussed in section~\ref{2metrics},
$\kappa$ quantifies the proportion of agreements among two coders that
are above chance. In Table~\ref{real-data}, both coders agree on 166
out of 242 coreference links (upper left cell).  Because of the
relatively high value of this cell, both recall and precision will be
high (cf. Figure~\ref{2by2schematic}).  But in addition, because the
proportion of coreference links is very high for both R$_{1}$
(179/242) and R$_{2}$ (185/242), the chance of agreement on
coreference links (or their absence) is also relatively high.
Factoring out this chance agreement results in poor reliability.

Table~\ref{rest-real-data} compares the $\kappa$ scores with recall
and precision for the same coder's annotations of ten narratives
from~\cite{chafe80} against the codings used
in~\cite{passonneau&litman97}.  Narrative one, with a $\kappa$ of .85
compared with recall and precision of .96, illustrates the general
trend that the $\kappa$ scores are good, but not as high as one might
assume given the generally high recall and precision. The last line of
the table gives the standard deviation ($\sigma$) for each metric.
Note that the standard deviation of the reliability measures is over 3
times that for recall and precision.  A log kept by the coder of
questions that arose during annotation suggests that the variation in
reliability reflects differences in the coherence of the narratives,
and the types of referential phenomena that occur, rather than
inconsistency in the coder's behavior.  For example, in this log the
coder reported greatest difficulty with narratives 9 ($\alpha$=.75)
and 12 ($\alpha$=.74), and used the phrases ``{\it I am confused, I
don't understand what he is talking about}'' to describe particular
coding problems.  In contrast, the coder described narrative 16
($\alpha$=.93) as ``{\it pretty easy to code}.''

{\small
\begin{table}
\begin{center}
\begin{tabular}{|c |r |r |r |}\hline
\multicolumn{1}{|c}{Narr.} &
\multicolumn{1}{|c}{$\kappa$} &
\multicolumn{1}{|c}{Recall} &
\multicolumn{1}{|c|}{Precision} \\\hline
 1 & .85 & .96 & .96 \\
 2 & .65 & .90 & .93 \\
 3 & .72 & .93 & .94 \\
 4 & .89 & .94 & .98 \\
 5 & .89 & .95 & .99 \\
 6 & .83 & .94 & .97 \\
 8 & .84 & .91 & .96 \\
 9 & .75 & .88 & .96 \\
11 & .79 & .92 & .95 \\
12 & .74 & .90 & .92 \\
15 & .80 & .93 & .93 \\
16 & .93 & .97 & .98 \\
17 & .86 & .95 & .96 \\
18 & .84 & .93 & .96 \\
19 & .85 & .96 & .93 \\
$\sigma$ & .07 & .02 & .02\\\hline
\end{tabular} 
\end{center}
\caption{\label{rest-real-data} Comparing Inter-rater Reliability of Coreference 
Annotations with Recall and Precision}
\end{table} }

\section{Summary}

A 2-by-2 coincidence matrix can be used to compute information
retrieval metrics, or to compute reliability.  Building on this
observation, I have shown how the method in Vilain et
al.~\shortcite{vilain&etal95} for computing recall and precision for
coreference annotation can be used to construct a coincidence matrix,
and therefore to compute reliability.  Each type of metric has its own
uses. If a target or correct annotation has been established, it may
be appropriate to evaluate recall and precision of a new coding
against the target.  However, in developing new annotated corpora with
no pre-exising answer key, so to speak, it is important to evaluate
the reliability of individual coders and of the datasets they produce.
The data presented in the preceding section
(Tables~\ref{real-data}-\ref{rest-real-data}) demonstrate that one
should not infer from high recall and precision of one annotation
against another that either annotation is reliable, in the sense of
reliability discussed in~\cite{cohen60} and~\cite{krippendorff80}.
Reliability measures should be used to identify reliable annotators
and annotations.  By merging the best data from mutually reliable
codings, a more correct coding can be derived for a new corpus.
Reliability scores can be used to determine whether a coder is
trainable (improvements over time), and when the training can be
terminated (no further improvement).  

Poor reliability can be an indicator of omissions or flaws in a coding
scheme.  In addition, reliability metrics can help the researcher
identify data that is consistently not agreed upon among multiple
coders.  This might occur within a single discourse for particular
kinds of coreference phenomena.  Or it might occur for an entire
discourse as compared with other discourses, e.g., if the discourse in
question is unclear, vague, or otherwise non-optimal for coreference
interpretation.

\section*{Acknowledgements}

Thanks to Pamela Jordan, Diane Litman, and Marilyn Walker for helpful
comments on this report.

\bibliographystyle{acl}

\end{document}